\documentclass[letterpaper, 10 pt, conference]{ieeeconf}  

\IEEEoverridecommandlockouts                              

\usepackage{graphicx}
\usepackage{makecell}
\usepackage{algorithm}
\usepackage{algpseudocode}

\title{\LARGE \bf
In-Context-Learning-Assisted Quality Assessment Vision-Language Models for Metal Additive Manufacturing
}

\author{Qiaojie Zheng, Jiucai Zhang, and Xiaoli Zhang$^{*}$
\thanks{$^{*}$Corresponding Author,
       Colorado School of Mines, Golden CO, USA
        {\tt\small xlzhang@mines.edu}}%
}

\begin{document}

\maketitle
\thispagestyle{empty}
\pagestyle{empty}

\begin{abstract}

Vision-based quality assessment in additive manufacturing often requires dedicated machine learning models and application-specific datasets. However, data collection and model training can be expensive and time-consuming. In this paper, we leverage vision-language models’ (VLMs’) reasoning capabilities to assess the quality of printed parts and introduce in-context learning (ICL) to provide VLMs with necessary application-specific knowledge and demonstration samples. This method eliminates the requirement for large application-specific datasets for training models. We explored different sampling strategies for ICL to search for the optimal configuration that makes use of limited samples. We evaluated these strategies on two VLMs, Gemini-2.5-flash and Gemma3:27b, with quality assessment tasks in wire-laser direct energy deposition processes. The results show that ICL-assisted VLMs can reach quality classification accuracies similar to those of traditional machine learning models while requiring only a minimal number of samples. In addition, unlike traditional classification models that lack transparency, VLMs can generate human-interpretable rationales to enhance trust. Since there are no metrics to evaluate their interpretability in manufacturing applications, we propose two metrics, knowledge relevance and rationale validity, to evaluate the quality of VLMs’ supporting rationales. Our results show that ICL-assisted VLMs can address application-specific tasks with limited data, achieving relatively high accuracy while also providing valid supporting rationales for improved decision transparency.
\end{abstract}

\section{Introduction}

uality assessment (QA) is a crucial step in ensuring part quality for additive manufacturing (AM). Traditional automated QA approaches based on machine learning (ML) or deep learning (DL) methods face two limitations that hinder their wide adoption. First, ML usually requires large, specialized datasets to train a task-specific model. The data collection process is expensive and time-consuming, and the trained models are usually not applicable beyond specific applications~\cite{Meng2020, Kumar2022, Trovato2025}. Second, ML approaches typically operate as black boxes that only output classification results without human-interpretable rationales to justify decisions. Such black-box-like behavior does not promote human trust, and therefore, these models are rarely deployed in real production~\cite{Linardatos2020}.

The reasoning capability of emerging large language models (LLMs) and vision-language models (VLMs) has demonstrated the potential to provide human-interpretable analysis in addition to classification results in applications such as medical X-ray analysis~\cite{Ryu2025, Tanno2024}. However, this has not been explored in QA for AM. To unleash its potential, two challenges need to be addressed. First, pretrained, off-the-shelf VLMs often lack domain-specific knowledge to make correct classification decisions and provide valid justification rationales~\cite{Urooj2025}. As metal AM is a highly specialized field where publicly available data is scarce, pretrained VLMs are not likely to acquire adequate knowledge~\cite{Fan2024}. Often, model fine-tuning is required to adjust model weights to equip them with appropriate knowledge and to align their attention mechanisms on how to examine AM prints. In addition, fine-tuned models typically remain limited to narrow use cases, and changes to the operating environment often require retraining. Second, metrics to quantify the soundness of the justification rationales produced by VLMs are lacking. Unlike the classification output, the supporting rationales are in natural language and therefore cannot be evaluated by simple metrics like classification accuracy.

\begin{figure}
\centering
\includegraphics[width=1.0\linewidth]{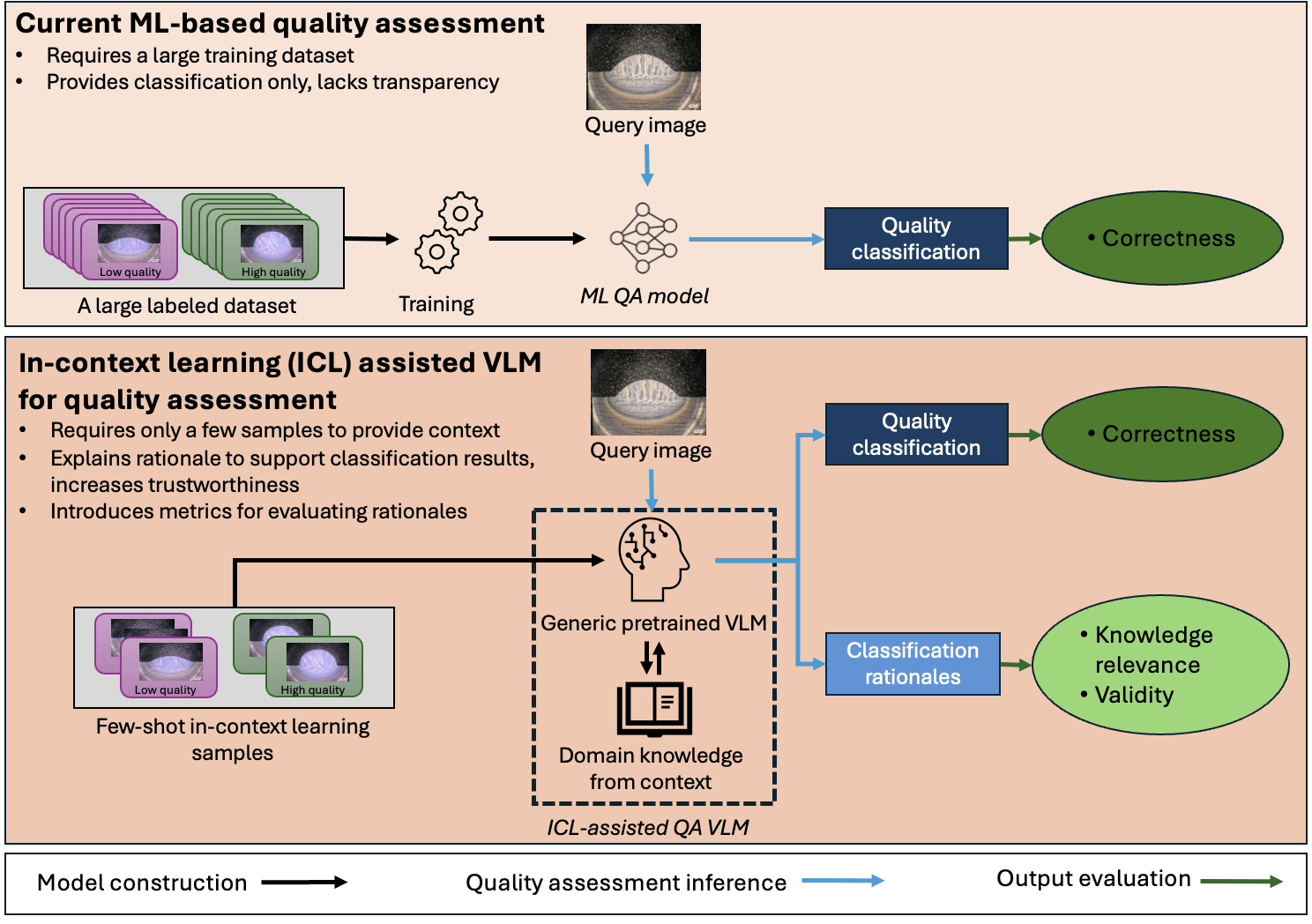}
\caption{Comparison of the existing ML-based QA framework with our VLM-based one. Our VLM-based framework does not require large, labeled datasets to construct an application-specific model; instead, it leverages VLMs' reasoning capabilities to learn the necessary QA knowledge from a few in-context samples. Compared with the ML-based approach, the VLM-based framework can produce a human-interpretable classification rationale to make QA decisions more trustworthy. We also present a set of metrics to evaluate the quality of the supporting rationale. }
\label{fig:Figure1}
\end{figure}

\begin{figure*}
    \centering
    \includegraphics[width=1.0\linewidth]{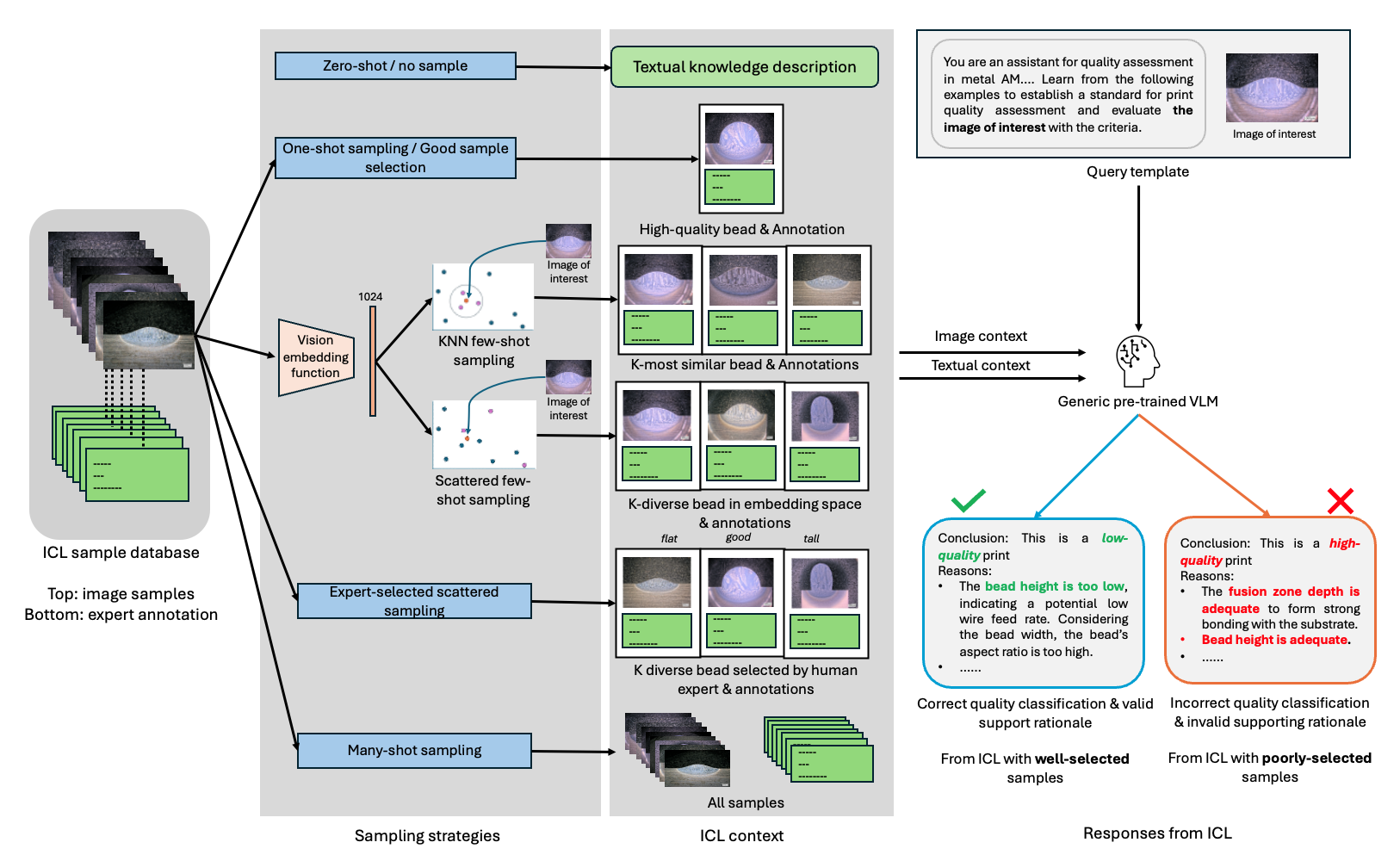}
    \caption{Different sampling strategies to prepare context for ICL. Six sampling strategies are explored in this study the influences of sampling strategy on quality assessment performance.}
    \label{fig:Figure2}
\end{figure*}

In this work, to enable VLM-based QA for AM, we present in-context learning (ICL) and a set of metrics to address the challenges of fine-tuning and the lack of evaluation metrics, respectively (shown in Figure~\ref{fig:Figure1}). First, ICL exploits the strong generalizability of VLMs and strategically uses a few demonstration samples that enable VLMs to quickly learn the key knowledge points and apply them to QA tasks. As a result, ICL does not update model weights, uses much less data, and is versatile to application or environment changes: replace the demonstration samples and VLMs can function under new settings. Second, we present a set of metrics that includes knowledge relevance, reasoning validity, and conclusion correctness to offer a quantitative evaluation of the output from ICL-assisted VLMs.

To further explore the best configuration for ICL, sampling strategies for context preparation for ICL were designed and evaluated, including variations in sample counts and selection criteria. QA experiments were conducted on various preparation strategies using two VLMs of different model sizes and reasoning capabilities, Gemma4-27b and Gemini 2.5 Flash, for AM QA, and the outputs were evaluated with the metrics presented above. Overall, the results demonstrate that VLMs of various sizes and capabilities can benefit from in-context learning, while the best-performing sampling strategies differ between models. In short, the contributions are as follows:

\begin{enumerate}
\item Designed different context preparation strategies for in-context learning and evaluated their performance.
\item Presented a VLM-based QA technique that not only provides an image classification result but also the reasoning behind it.
\item Provided evaluation metrics to measure the performance of VLM-based QA.
\item Validated the performance in physical settings with data collected from metal AM processes.
\end{enumerate}

\section{Related works}

\subsection{Quality assessment in additive manufacturing}
Recent advancements in DL, particularly convolutional neural networks (CNNs), have driven significant progress in vision-based QA and defect detection for AM. For example, Cui et al.~\cite{Cui2020} presented a CNN model to classify defects in metal AM, achieving 92.1\% on optimal microscope images. Similar CNN-based approaches, including object detection frameworks like YOLO and region-based CNNs, have been applied to scanning electron microscopy, optical, and in-situ monitoring images with strong performance~\cite{Wen2021, Yin2025, Wang2024}. However, these ML-based approaches face two major limitations: demanding training-data requirements and lack of interpretability, as discussed in the Introduction.

\subsection{VLM/LLM applications in additive manufacturing}

A handful of recent attempts have investigated LLMs/VLMs for AM applications~\cite{Chandrasekhar2024, Fan2024}. Because of their natural language processing capabilities, these models can respond to human queries in a human-interpretable manner, reducing the need for users to possess deep technical expertise when understanding AM-related issues. Despite their promise, the primary challenge in applying LLMs/VLMs lies in their limited domain-specific knowledge. This challenge is especially noticeable in the metal AM domain, where publicly available training data is scarce for pretraining. To bridge this gap, AMGPT ~\cite{Chandrasekhar2024} employed the Retrieval-Augmented Generation (RAG) method~\cite{RAG_PAPER} for LLM-based material discovery. However, this approach has notable limitations: RAG is highly effective for text-centric tasks but struggles to capture the multimodal information required for metal AM.

\subsection{In-context learning}
In-context learning offers an efficient alternative to fine-tuning for incorporating application-specific knowledge. By providing only a few labeled examples in the prompt, LLMs can rapidly infer the necessary knowledge and generalize it to application-specific inputs. In-context learning has been adopted and shown to be effective for a variety of text-based tasks, such as dialogue state tracking, semantic parsing, and cross-domain question answering~\cite{Hu2022, An2023, Mei2025}.
Despite its prominence in natural language processing, its application to image-based tasks remains underexplored. Prior work is limited to a handful of medical image classification studies~\cite{Ferber2024}; to our knowledge, there are no reports of its use for QA in AM settings. Most existing efforts focus exclusively on classification, leaving open scenarios where VLMs must not only classify images accurately but also provide domain-specific, human-interpretable rationales for their predictions.

\section{Methodology}
\subsection{Sampling for in-context learning}
The effectiveness of in-context learning is highly dependent on the sampling strategy. In this work, ICL strategies are systematically designed and evaluated, focusing on two factors: the number of samples and the diversity of samples. Building on these two factors, six sampling strategies are designed to assess their influence on model performance, as shown in Figure~\ref{fig:Figure2}. Each image sample is labeled with accompanying human-expert annotations. These samples are ordered by their similarity to the images of interest in the embedding space~\cite{Guo2024}.

\subsubsection{No-sample/Zero-shot}
In zero-shot sampling, ICL contains only essential textual knowledge relevant to quality evaluation, without including any example images or expert-provided annotations. This setup enables assessment of the model’s inherent capability to perform QA based solely on general instruction, without leveraging prior exposure to domain-specific demonstrations.

\subsubsection{One-shot sampling}

One-shot ICL includes a manually selected sample that contains a high-quality bead image. The idea in this sampling strategy is to let VLMs infer key characteristics that define a high-quality bead and use this as a reference standard. At test time, an input that does not meet this standard can be classified as low quality and supported by corresponding reasoning on subpar characteristics. Since this sample is manually selected by a human expert, it does not change throughout testing.

\subsubsection{K-nearest-neighbor (KNN) few-shot sampling}
The KNN few-shot sampling automatically selects a small set of samples from the database whose images are most similar to the query image in the embedding space. The idea behind this approach is to let VLMs learn and generalize the necessary knowledge from a few samples that are similar to the query image. In this study, image embeddings are generated using a CLIP model~\cite{Radford2021}, which converts each image into a 1024-dimensional feature vector. Similarity between images is then measured by cosine similarity (Equation~\ref{eq:cosine_similarity}) to retrieve the most similar samples. $\mathbf{im_{eval}}$ is the image of the bead to be evaluated, and $\mathbf{im_{database}}$ is the image from the database. Samples used in the KNN sampling strategy change depending on the input during testing.

\begin{equation}
\label{eq:cosine_similarity}
\mathrm{similarity} =
\frac{\mathbf{im_{eval}} \cdot \mathbf{im_{database}}}{\|\mathbf{im_{eval}}\| \, \|\mathbf{im_{database}}\|}
\end{equation}

\subsubsection{Scattered few-shot sampling}
The scattered few-shot sampling strategy aims to create in-context learning prompts that contain a diverse set of samples, providing wide coverage of QA knowledge. This strategy automatically selects samples that are as widely spread as possible in the embedding space, letting the model build a robust knowledge foundation for evaluating unseen inputs. For instance, when $k=3$, the selected examples consist of the most similar sample, the most dissimilar sample, and one that lies near the middle of the distribution.

\subsubsection{Expert-selected scattered few-shot sampling}
Similar to the scattered few-shot sampling approach, the aim of this one is also to provide wide coverage, but from the perspective of expert understanding of QA knowledge rather than embedding distances. Human experts select samples based on established knowledge about bead classes and their failure reasons. For example, when selecting three samples for the prompt, the expert might include one high-quality bead, one low-quality bead with insufficient height, and another low-quality bead with excessive height. By deliberately covering a wide range of defect types, the expert-selected samples expose VLMs to diverse failure modes and provide rich, domain-specific knowledge, thereby enhancing its ability to assess bead quality accurately.

\subsubsection{Many-shot sampling}
Many-shot sampling uses all available samples in the database. The goal is to maximize the information provided to VLMs, exposing it to the full range of bead qualities and defect types. By leveraging the entire dataset, this approach aims to enable VLMs to build a more comprehensive understanding of quality standards and improve its ability to generalize to unseen inputs.

\subsubsection{VLM Baseline}
A QA task query that contains no knowledge serves as the explanation and classification baseline for the sampling strategies discussed above.

\subsection{In-context learning QA query structure}
The overall query structure is presented in Algorithm~\ref{algo:algo1}. The user supplies the ICL context database composed of images and the corresponding expert annotations. The user may also define the sampling function to form the ICL context. A system-level instruction is predefined by expert to guide pretrained VLMs to leverage the ICL context to perform QA. The user-selected VLM then combined all these information to perform QA on the image of interest.

\begin{algorithm}
    \caption{ICL-assisted VLM Quality Assessment}\label{algo:algo1}
    \begin{algorithmic}[1]
    \Require Image to be evaluated $\mathrm{Im_{eval}}$
    \Require ICL image sample database $D_{im}$
    \Require ICL image annotation database $D_{an}$
    \Require VLM to perform QA task $VLM$
    \Require Vision embedding model to embed image $EMB$
    \Require Sampling function $SMP$
    \Procedure{VLM\_QA}{$\mathrm{Im_{eval}}$, $D$, $VLM$, $EMB$}
    \State Calculate embeddings for image database $E \gets EMB(D_{im})$
    \State Calculate embedding for image to be evaluated $e \gets EMB(\mathrm{Im_{eval}})$
    \State Get desired sample index $i\gets SMP(e, E)$
    \State Collect images samples for ICL $\mathrm{Im_{ICL}} \gets D_{im}[i]$
    \State Get annotation samples for ICL $\mathrm{An_{ICL}} \gets D_{an}[i]$
    \State Expert-defined task instruction $inst$
    \State Get QA output with $QA_{output} \gets VLM(\mathrm{Im_{ICL}, \mathrm{An_{ICL}}, \mathrm{Im_{eval}}}, inst)$
    \EndProcedure
\end{algorithmic}    
\end{algorithm}

\subsection{Evaluation metrics}
We define three metrics to evaluate the QA outputs from VLMs from two perspectives: rationale (metrics 1 and 2)and classification correctness (metric 3). These three metrics are presented in detail below.
\begin{enumerate}
\item Knowledge relevance assesses whether the model focuses on the correct regions and applies domain-specific knowledge to justify its quality classification in the rationale. It is scored proportionally to the number of relevant knowledge points covered, with deductions for inappropriate knowledge, ranging from 0 to 1.

\item Rationale validity evaluates whether the model’s rationale is logically sound, correctly identifies relevant image features, and appropriately interprets them to support its quality-classification decision. This is a binary metric assessed by human experts.

\item Conclusion correctness solely focuses on the accuracy of the classification decision. This metric provides a measure comparable to the accuracy from traditional ML methods for performance comparison. 

\end{enumerate}

Note that knowledge relevance and rationale validity provide evaluations of the supporting rationales from distinct perspectives. Knowledge relevance evaluates whether the knowledge applied in QA by VLMs is appropriate and sufficient for the specific application, and rationale validity measures whether the knowledge is used correctly by VLMs to assess the quality of the print. High knowledge relevance is a prerequisite for high rationale validity, but it does not guarantee high rationale validity.


\section{Experiment Setup}
\subsection{Printing setup}

\begin{figure}
    \centering
    \includegraphics[width=1.0\linewidth]{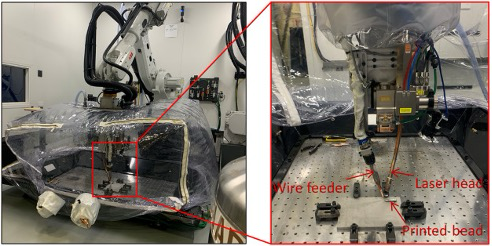}
    \caption{Laser-wire direct energy deposition setup used to manufacture all learning and testing samples. The laser head is mounted on an industrial robot. Image from~\cite{Liu2022}}
    \label{fig:Figure3}
\end{figure}

The experimental samples were manufactured using a wire-laser direct energy deposition process with Ti-6Al-4V, following the setup shown in Figure~\ref{fig:Figure3}. A 6kW laser head was used to melt the wire, which was continuously supplied by a precision wire feeder. The laser head and wire feeder were mounted on an industrial robot arm to ensure precise control of the deposition path. The process was carried out in a sealed chamber filled with argon gas to maintain an inert environment and prevent oxidation. Various combinations of wire feed rate, laser power, and robot travel speed were used to manufacture tracks of different qualities.

\subsection{Data composition} A total of 65 single-bead samples were manufactured. Of these, 9 representative samples were manually annotated with both a quality classification and supporting reasons, and the remaining samples were used for testing. The 9 annotated samples were used to construct a database from which in-context learning prompts were drawn. Two defect categories were included in the selected samples: excessive bead height and insufficient bead height. For the few-shot testing cases, we set the size of $k$ to 3, as this is the minimum number required to cover all cases. While testing with varying $k$ would yield a more comprehensive evaluation, this setup would generate too many results to evaluate manually.

\begin{figure}
    \centering
    \includegraphics[width=1.0\linewidth]{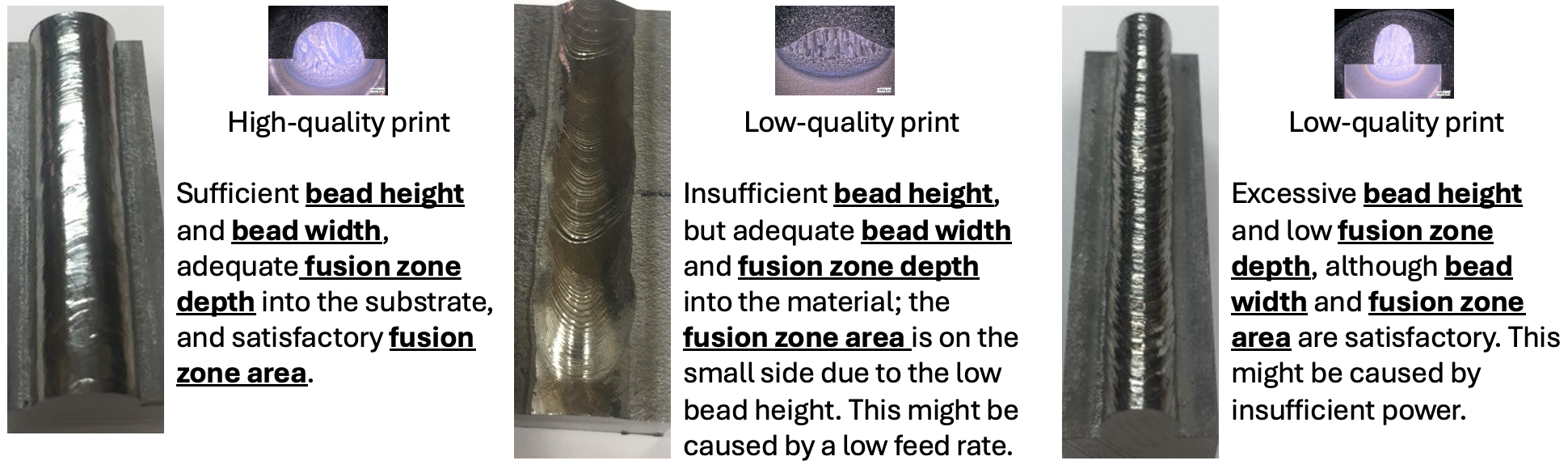}
    \caption{Print quality visualization and expert comments of selected samples. The bold, underscored texts highlight the knowledge points should be used when evaluating the quality. Low-quality bead experiences non-smooth surface finishes. Note that only cross-sectional images are used in QA query and ICL. The surface finish images are presented for visual reference of how bead quality is defined.  }
    \label{fig:Figure4}
\end{figure}

\begin{figure*}
    \centering
    \includegraphics[width=1.0\linewidth]{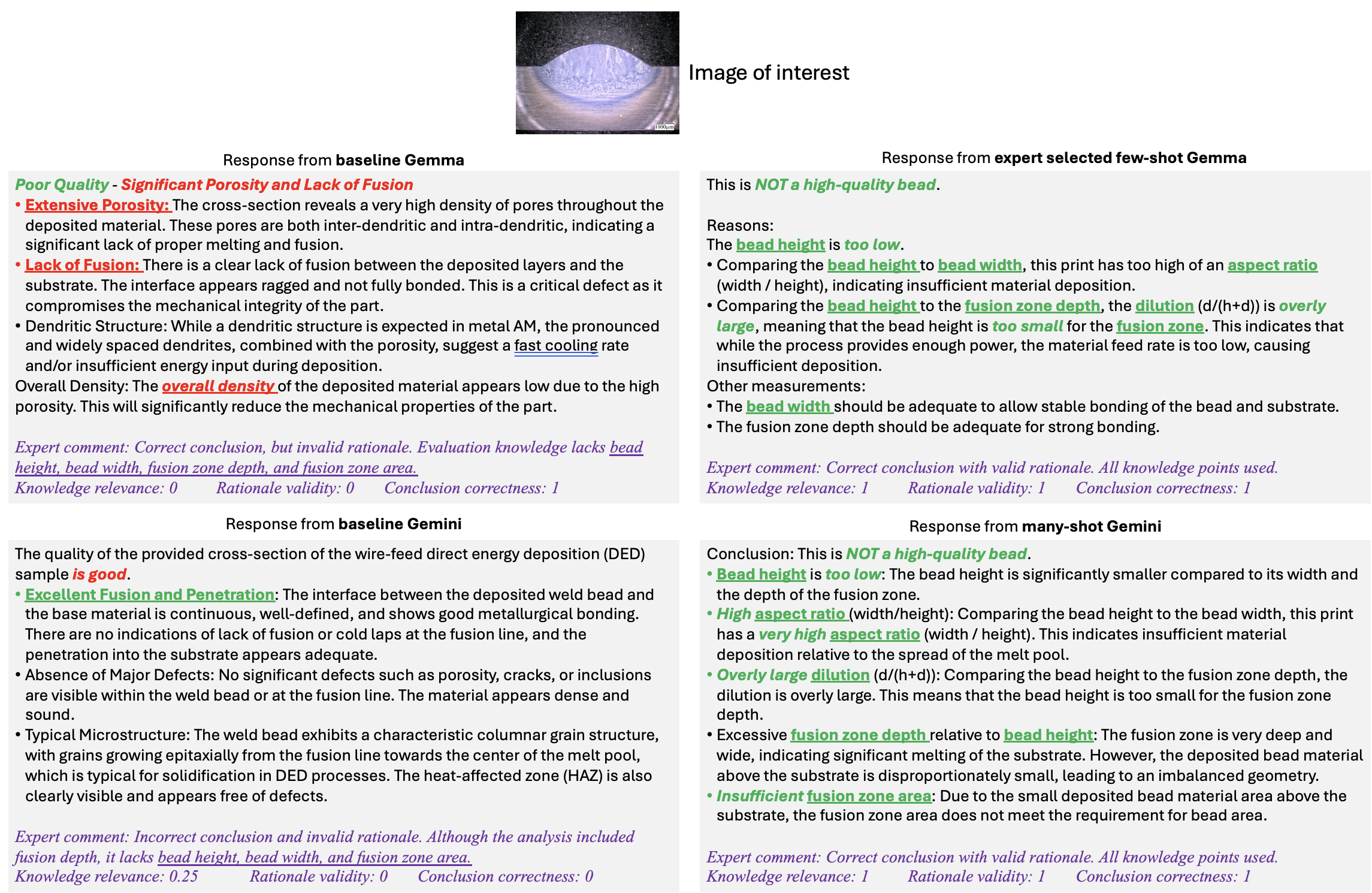}
    \caption{Sample outputs from VLMs for QA on the image of interest. The selected image is a borderline low-quality case to provide a challenging evaluation of ICL effectiveness. The top row contains responses from Gemma, and the bottom row contains responses from Gemini. The left column contains baseline responses from both models, and the right column contains responses from the best ICL configurations for both models. Red bold italic text indicates incorrect conclusions, while red bold underlined text indicates incorrect knowledge points; green marks their correct counterparts. Purple text shows comments on the QA output from human experts. Underlined purple text indicates comments on knowledge points. }
    \label{fig:Figure5}
\end{figure*}

\subsection{Data annotation}
The 9 images in the database were annotated by human experts to provide ICL context to VLMs. It is structured in the following three parts: 1) overall print quality, whether it is a high-quality bead or not; 2) comments on the unsatisfactory feature if the current one is a low-quality bead; and 3) comments on the satisfactory feature.
Three examples of the produced beads with abbreviated human-expert comments are shown in Figure~\ref{fig:Figure4}. The two right images show ripped and discontinuous beads, which are low-quality samples; the left one is a high-quality bead with a smooth finish. Images used in queries and ICL only contains the cross-sectional images, surface is presented for readers' visual references of bead quality definition. 

\subsection{Domain knowledge in quality assessment}
The domain knowledge in this study involves four measurements of the bead cross section, bead height, bead width, fusion zone depth, and fusion zone area, and their derived quantities, including aspect ratio and dilution. A smooth bead is correlated with a dome-shaped bead which poses certain requirement on the bead height, and aspect ratio with the bead width. For sufficient bonding with the substrate, fusion zone must penetrate deep enough, and this poses requirements on the fusion zone depth and its proportion to the bead height. Finally, to ensure there is enough material being deposited, a requirement on fusion zone area is also posted. Therefore, to accurately assess the quality of the print, VLMs must possess the described knowledge based on the four fundamental dimensional measurements of the cross sectional geometry. Due to the inherent inaccuracies of VLMs when performing numerical measurements~\cite{Yang2024}, the knowledge presented in ICL is qualitative. VLMs need to generalize the good-bead standard by interpreting the knowledge descriptions and annotations by correlating them with their perception of cross sectional images. 


\subsection{VLM selection}
The capability of the backbone VLMs is also a critical factor influencing performance. To ensure a fair evaluation of the in-context learning strategy, we tested its performance using two VLMs of different sizes and capabilities: Gemini 2.5 flash and the 27-billion-parameter Gemma 3. Gemini 2.5 flash is a large-scale model with higher reasoning capability; this model is accessed through a cloud API, while Gemma 3 is a smaller model hosted locally. These two models allow us to compare the effectiveness of the in-context learning strategy on VLMs with distinct reasoning characteristics.

\section{Results and discussions}
Knowledge relevance, reasoning validity, and conclusion correctness across 6 sampling strategies and the baseline are summarized and presented as a radar plot in Figure~\ref{fig:Figure5}. The left panel shows the evaluation performed on Gemma, and the right panel shows the evaluation on Gemini. To provide a detailed explanation of the performance, we also include selected responses in Figure~\ref{fig:Figure6}. The code and full testing results will be provided through our github after the anonymous review process.

\subsection{Metrics effectiveness}
We first examine the proposed metrics with examples presented in Figure~\ref{fig:Figure5}. The left column contains QA responses from generic pretrained models, and the right column contains those from their best ICL configurations.

The knowledge relevance metric accurately reflects the lack of application-specific knowledge in both explainability baseline models. The supporting rationales do not cover all the geometry-based evaluations required in this application. For the Gemma model, it even focused on the background supporting material and falsely recognized the white sparkles as a lack of fusion. For the ones assisted by ICL, knowledge relevance reflects the improvements, as the supporting rationales are based on geometric measurements.

Rationale validity reflects the inappropriateness and incompleteness of the rationales in the two VLMs baselines. Supporting rationales from Gemma are deemed invalid since its responses are not based on the correct application-specific reasons. For the baseline Gemini model, even though it evaluates the bead with portions of application-specific knowledge, the model still misses critical evaluations in bead height and bead area, making the assessment invalid. With ICL, rationale validity can reflect the correct application of the application-specific knowledge that helped identify the quality problem.

Finally, conclusion correctness reflects an evaluation of quality classification detached from the supporting rationales. In the baseline Gemma case, the conclusion correctness is deemed correct, while the rationale is unsatisfactory.

\begin{table*}

\centering

\caption{Performance Summary of Different Sampling Strategies}
\begin{tabular}{|c|c|c|c|c|c|c|}
\hline
& \multicolumn{3}{c|}{Gemini 2.5 Flash} & \multicolumn{3}{c|}{Gemma3:27b}\\
\hline
Sample strategy & \makecell{Conclusion\\correctness} & \makecell{Rationale\\validity} & \makecell{Knowledge\\relevance}& \makecell{Conclusion\\correctness} & \makecell{Rationale\\validity} & \makecell{Knowledge\\relevance} \\
\hline
VLM Baseline &0.37 & 0.38 & 0.3 & 0.64& 0 & 0 \\
\hline
Zero-shot & 0.57 & 0.55 & 0.99& 0.72& 0.17 & 1 \\
\hline 
One-shot & 0.56 & 0.52 & 0.98& 0.37& 0.36 & 0.98  \\
\hline
KNN few-shot & 0.61 & 0.57 & 0.99& 0.49& 0.43 & 1 \\
\hline 
Scattered few-shot & 0.7 & 0.7 & 1& \textbf{0.7}& \textbf{0.58}&\textbf{1} \\
\hline
Expert selected few-shot & 0.64 & 0.64 & 0.99& 0.64& 0.57 &1\\
\hline
Many-shot & \textbf{0.75} & \textbf{0.75} & \textbf{0.99}& 0.45& 0.34 & 1 \\
\hline
\multicolumn{7}{c}{ Conclusion correctness from ML-based quality assessment on similar dataset is 0.75\cite{Liu2022}.} \\

\end{tabular}
\label{tab:table1}
\centering
\end{table*}

\subsection{Influences of sampling strategy}

\subsubsection{Influence on knowledge relevance}
\begin{figure}
    \centering
    \includegraphics[width=1.0\linewidth]{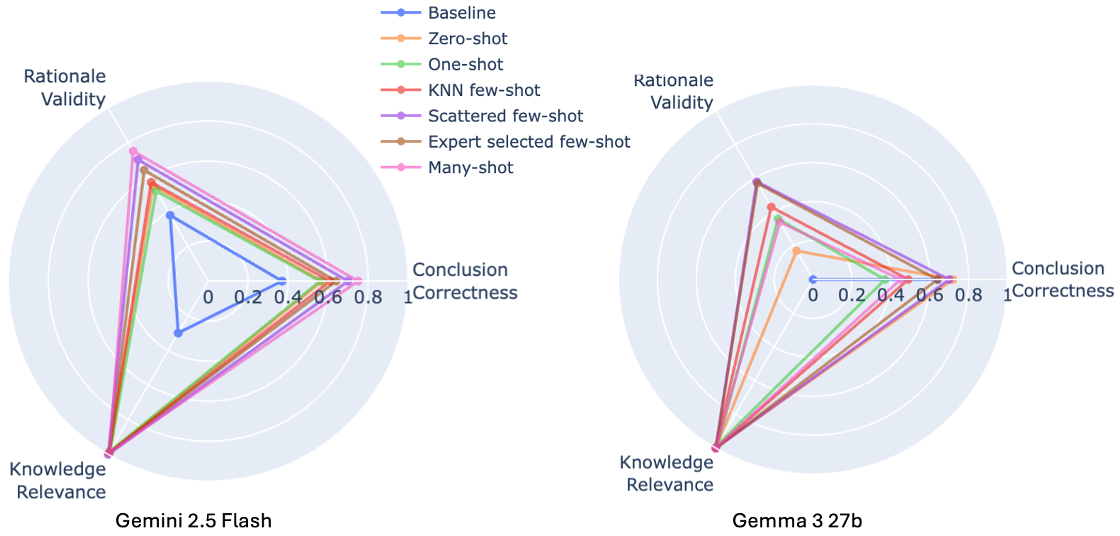}
\caption{Spider plot comparing assessment quality (rationale validity, knowledge relevance, and conclusion correctness) across ICL sampling strategies. All ICL strategies outperform the baseline query for both VLMs, with many-shot ICL yielding the highest quality in Gemini and expert-selected few-shot sampling performing best in Gemma. Variations in assessment quality from Gemini are smaller than those from Gemma, indicating that Gemini is less sensitive to ICL samples than Gemma is. Overall, Gemini achieves higher assessment quality than Gemma across all ICL cases and the baseline, illustrating its stronger reasoning capability and knowledge foundation.}
\label{fig:Figure6}
\end{figure}

Knowledge relevance is not greatly affected by sampling strategies, as shown by the relatively concentrated data points on the knowledge relevance axis in Figure~\ref{fig:Figure6} and the near-1 values in the corresponding columns in Table~\ref{tab:table1}. This indicates that, even with only textual knowledge descriptions, VLMs can reliably acquire application-specific knowledge and apply it to domain-specific tasks.
\subsubsection{Influence on rationale validity}
Rationale validity is greatly influenced by the sampling strategies, as correctly understanding, generalizing, and applying knowledge to unseen images is a much more challenging task. The two VLMs also display different preferences for sampling strategies.

Gemini shows a higher preference for many-shot sampling strategies that include a large number of diverse samples, while Gemma prefers few-shot strategies that offer a balanced context length. This preference may be explained by the difference in reasoning capability between the two models: higher reasoning capability allows larger VLMs to filter out context noise in rich, long contexts and to generalize application-specific knowledge, whereas the smaller model is likely overwhelmed by the long context and therefore benefits from sampling strategies that offer a balanced and compact context.

In terms of sampling diversity, both models prefer diversity-focused strategies over similarity-focused strategies. As demonstrated in the three few-shot configurations, the scattered few-shot and expert-selected few-shot approaches achieve similar validity scores within testing performed using the same model, and both are higher than those from the KNN approach. Diversity-focused context, therefore, is more likely to promote VLMs to learn generalizable knowledge, regardless of VLM size and capability.

With a further reduction in the number of samples included, rationale validity decreases, as demonstrated by the zero- and one-shot cases. It is important to note that the larger Gemini model is less affected by this reduction than the Gemma model, indicating stronger generalizability.
\begin{figure*}
    \centering
    \includegraphics[width=1.0\linewidth]{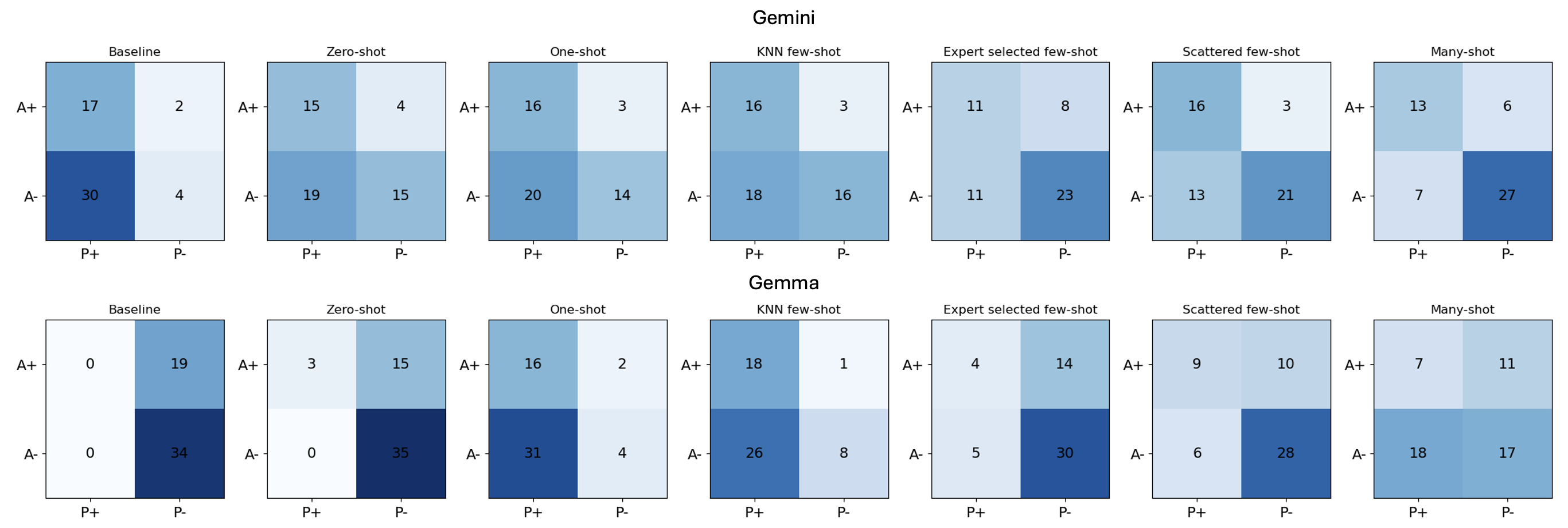}
    \caption{Confusion matrix of quality classification across various sampling strategies. $P+/P-$ and $A+/A-$ indicate predicted and actual high/low-quality beads, respectively. Although the models displays different biases in the baseline, ICL can effectively corrects the bias. Gemini performed best with many-shot sampling, while Gemma performed best with scattered few-shot sampling, which displays consistent pattern as observed in validity evaluation.}
    \label{fig:ConfusionMatrix}
\end{figure*}

\begin{figure*}
    \centering
    \includegraphics[width=1.0\linewidth]{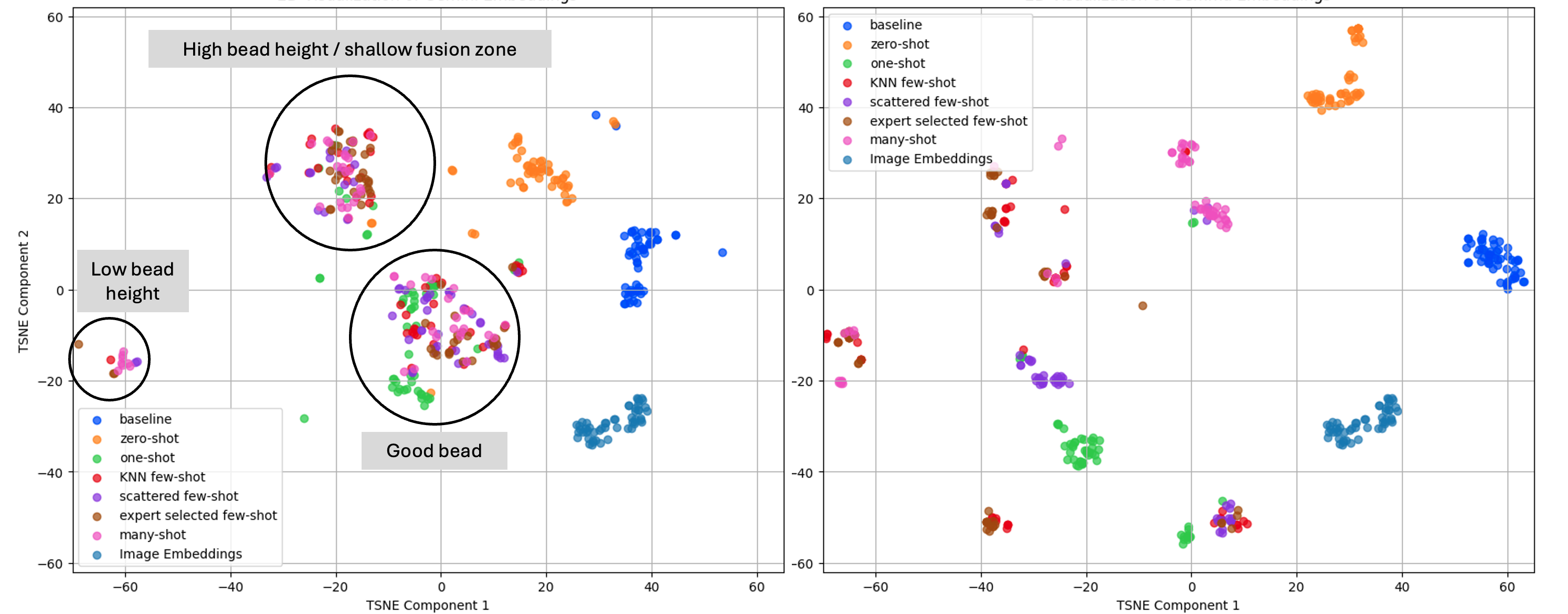}
    \caption{Semantic distributions of QA responses from Gemini and Gemma under different ICL configurations. Responses were embedded using the OpenCLIP model and reduced to two dimensions with t-SNE. Gemini’s responses form large overlapping regions across ICL configurations, indicating low sensitivity to context variation and more consistent generalization. The circled overlapping regions are also consistent with expert observations of bead quality in the data. In contrast, Gemma’s responses are more scattered, reflecting higher sensitivity to context and greater variability in reasoning. It should be noted that the circles are manual estimates and do not indicate hard distribution boundaries. This distributional difference illustrates the stronger reasoning capacity of Gemini, which enables it to generalize from context more effectively, while Gemma remains more influenced by the input configuration.}
\label{fig:Figure7}
\end{figure*}

\subsubsection{Influence on conclusion correctness} The influence on conclusion correctness is largely the same, as logical rationales naturally promote the correct conclusions. To visualize the detailed classification behaviors, the confusion matrices across all sampling strategies were shown in Figure~\ref{fig:ConfusionMatrix}. Even though these two models have different classification bias when used as-is, ICL can effectively correct their classification bias. The overall trend matches those observed in validity, that is Gemini reached the best performance under the many-shot setting, while the smaller Gemma model reached the best performance under scattered few-shot. It is interesting to note that Gemma's prediction under the many-shot setting almost reaches an even split, this is likely caused by the overwhelming long context that the smaller model cannot process correctly. 


\subsection{Influence of model sizes}
Larger models achieve overall higher rationale validity and conclusion correctness compared to the smaller model, as illustrated by the overall outward distribution shown in Figure~\ref{fig:Figure6}. This is because the stronger reasoning capability lets the larger model generalize knowledge in ICL better. 

Additionally, larger model exhibits smaller changes in responses under various sampling strategies, showing less sensitivity to the changes in context. To visualize such conclusion, the dimensionality-reduced response embeddings is visualized in Figure~\ref{fig:Figure7}. The evaluation responses are first converted into embedding vectors with the OpenCLIP model, which is the same one used to find image similarity. t-SNE is then used to reduce the dimension from 1024 to 2 to allow 2D plotting. Gemini's response forms overlapping regions across different sampling strategies while those from Gemma is more scattered apart. It is interesting to note that the clustering formed by Gemini can be grouped by the types of beads in the dataset. The high overlapping in Gemini's response concludes that a larger model is less sensitive to context change.  


\section{Conclusion}
This work applies ICL with VLMs for QA tasks in metal AM to avoid the large dataset requirements needed in traditional ML approaches and to provide quality-classification rationales for trustworthiness. Sampling strategies were studied to generate ICL content, and new evaluation metrics, knowledge relevance, rationale validity, and conclusion correctness were introduced to evaluate the quality of ICL-assisted QA. Evaluations with two VLMs of different sizes and capabilities (Gemma-3:27b and Gemini-2.5-Flash) showed that both can achieve classification accuracy comparable to traditional ML methods. Both models experienced noticeable improvements from ICL compared to their bare-bones, pretrained counterparts, highlighting the effectiveness of ICL. With extensive evaluations across various sampling strategies, we found that the larger model (Gemini) benefits from rich context with numerous diverse samples, while the smaller model (Gemma) benefits from a balanced-length context composed of diverse samples.

\bibliographystyle{IEEEtran}
\bibliography{ICL}

\end{document}